\pdfoutput=1

\documentclass[11pt]{article}

\usepackage{naacl2021}
\usepackage{examples}
\usepackage{times}
\usepackage{latexsym}

\usepackage[english]{babel}
\usepackage[utf8x]{inputenc}
\usepackage{amsmath}
\usepackage{graphicx}
\usepackage{multirow}
\usepackage{url}
\usepackage{xspace}
\usepackage{booktabs}
\usepackage[colorinlistoftodos]{todonotes}
\usepackage{times}
\usepackage{amsthm}
\usepackage{latexsym}
\usepackage{graphicx}
\usepackage{mathtools}
\usepackage{amsmath}
\usepackage{amssymb}
\usepackage{tikzscale}
\usepackage[export]{adjustbox}
\usepackage{arydshln}
\usepackage{multirow}
\usepackage{rotating}
\usepackage{csquotes}
\usepackage{pifont}
\usepackage[T1]{fontenc}
\usepackage{subcaption}
\usepackage[first=0,last=9]{lcg}
\usepackage{color, colortbl}
\definecolor{Gray}{gray}{0.9}
\definecolor{LightCyan}{rgb}{0.90,1,1}

\usepackage{microtype}

\newcommand{\q}[1]{``#1''}
\newcommand{\specialcell}[2][c]{%
  \begin{tabular}[#1]{@{}l@{}}#2\end{tabular}}

\title{Probing for Bridging Inference in Transformer Language Models}

\author{Onkar Pandit \\
  University of Lille, INRIA Lille, \\
  CNRS, Centrale Lille,UMR 9189-CRIStAL,\\ F-59000, Lille, France \\
  \texttt{onkar.pandit@inria.fr} \\\And
  Yufang Hou \\
  IBM Research Europe \\
  Dublin, Ireland \\
  \texttt{yhou@ie.ibm.com} \\}

\begin{document}
\maketitle
\begin{abstract}
We probe pre-trained transformer language models for bridging inference. We first investigate individual attention heads in BERT and observe that attention heads at higher layers prominently focus on bridging relations in-comparison with the lower and middle layers, also, few specific attention heads concentrate consistently on bridging. More importantly, we consider language models as a whole in our second approach where bridging anaphora resolution is formulated as a masked token prediction task (\emph{Of-Cloze test}). Our formulation produces optimistic results without any fine-tuning, which indicates that pre-trained language models substantially capture bridging inference. 
Our further investigation shows that the distance between anaphor-antecedent and the context provided to language models play an important role in the inference.
\end{abstract}

\section{Introduction}
Bridging inference involves connecting conceptually 
related discourse entities $-$ anaphors and antecedents \cite{clarkherberth75}. A bridging anaphor shares \emph{non-identical} relation with its antecedent and depends on it for complete interpretation. This 
differs from \emph{coreference resolution} which links 
mentions 
that refer to the same entity (i.e., mentions in the same entity share identical relations). Consider the following example $-$

\q{\textit{In \underline{Poland's} rapid shift from socialism to an undefined alternative, environmental issues have become a cutting edge of broader movements to restructure \textbf{the economy}, cut cumbersome bureaucracies , and democratize local politics.}}

Bridging inference connects the anaphor ``\textbf{\emph{the economy}}'' and its antecedent ``\underline{\emph{Poland}}'' and deduces that \q{the economy} specifically refers to \q{the economy of Poland}.

We want to investigate if the pre-trained transformer language models capture 
any 
bridging inference information. Recently there has been an increasing interest in analyzing pre-trained language models' ability at capturing syntactic information \cite{clark-etal-2019-bert}, semantic information \cite{kovaleva-etal-2019-revealing},  as well as commonsense knowledge \cite{talmor2019olmpics}. There are also a few studies focusing on probing coreference information in pre-tained language models \cite{clark-etal-2019-bert, sorodoc-etal-2020-probing}. So far, there has no work on analyzing bridging, which is an important type of entity referential information. We try to fill this gap in our work.

We employ two different but complementary approaches for the probing of pre-trained transformer language models for bridging inference. In the first approach (Section \ref{sec:head}), we investigate the core internal part of transformer models -- self-attention heads in vanilla BERT \cite{devlin-etal-2019-bert}. We look at the attention heads of each layer separately and measure the proportion of attention paid from anaphor to antecedent and vice versa. This captures the magnitude of bridging signal corresponding to each attention head. We observed that attention heads of higher layers are more active at attending at bridging relations as well as some of the individual attention heads prominently look at the bridging inference information. 

In the second approach (Section \ref{sec:fill-in-the-gap}), we treat pre-trained transformer language models as a black box and form bridging inference as a masked token prediction task. This formulation takes into consideration the whole architecture and weights of the model rather than concentrating on individual layers or attention heads, thus, complementing our first approach where we looked at the individual parts of the transformer model. For each bridging anaphor, we provide input as \q{\emph{context anaphor of [MASK]}} to language models and get the scores 
of different 
antecedent candidates
for mask tokens. 
We then
select the highest scoring candidate as the predicted antecedent. Surprisingly, the best variation of this approach produces a high accuracy score of 28.05\% for bridging anaphora resolution on ISNotes \cite{markert-etal-2012-collective} data without any task-specific fine-tuning of the model. On the same corpus, the current state-of-the-art bridging anaphora resolution model \emph{BARQA} \cite{hou-2020-bridging} 
achieves an accuracy of 50.08\%, while a solid mention-entity pairwise model with carefully crafted semantic features \cite{hou-etal-2013-global} produces an accuracy score of 36.35\%.  This shows that substantial bridging information is captured in the pre-trained transformer language models.

Bridging inference requires both commonsense world knowledge as well as context-dependent text understanding. The above-mentioned fill-in-the-gap formulation for the antecedent selection task is flexible enough to easily explore the role of different types of context for bridging inference.  
 Our analysis shows that pre-trained language models capture bridging inference 
 substantially
 however the overall performance depends on the context provided to the model. It is also observed that bigger language models are more 
accurate
at capturing bridging information.

This work has two main contributions. First, we thoroughly investigate bridging information encoded in pre-trained language models using two probing approaches (\emph{attention heads analysis} and \emph{fill-in-the-gap}). Second, we provide a deeper understanding of the bridging referential capabilities in the current pre-trained language models. Our experimental code is available at \url{https://github.com/oapandit/probBertForbridging}.

\section{Related Work}

\paragraph{Entity Referential Probing.} Previous studies on entity referential probing mainly focus on coreference. \citet{clark-etal-2019-bert} showed that certain attention heads in pre-trained BERT correspond well to the linguistic knowledge of coreference. Particularly, the authors found that one of BERT’s attention heads achieves reasonable coreference resolution performance compared to a string-matching  baseline  and  performs  close  to  a simple rule-based system. \citet{sorodoc-etal-2020-probing} investigated the factors affecting pronoun resolution in transformer architectures. They found that transformer-based language models capture both grammatical properties and  semantico-referential information for pronoun resolution. 
Recently, \citet{hou2020finegrained} analyzed the attention patterns of a fine-tuned BERT model for information status (IS) classification and found that the model pays more attention to signals that
correspond well to the linguistic features of each IS class. For instance, the model learns to focus on a
few premodifiers (e.g., ``more'', ``other'', and ``higher'') that indicate the comparison between two entities. In this work, we focus on probing bridging, which is a more challenging entity referential relation and one of the oldest topics in computational linguistics \cite{clarkherberth75,bos95,asher98}. 

\paragraph{Attention Analysis.}
Recently there has been an increasing interest in analyzing attention heads in transformer language models. Although 
some researchers argue that attention does not explain model predictions \cite{jain-wallace-2019-attention}, analyzing attention weights still can help us to understand information learned by the models \cite{clark-etal-2019-bert}.
Researchers have found that some BERT heads specialize
in certain types of syntactic relations \cite{Htut2019DoAH}.
\citet{kovaleva-etal-2019-revealing} reported that pre-trained BERT’s heads encode information correlated to FrameNet’s relations between frame-evoking lexical units (predicates, such as ``\emph{address}'') and core frame  elements (such as ``\emph{issues}'').
 In our work, we try to analyze whether certain attention heads in a pre-trained BERT model capture bridging relations between entities in an input text. 


\paragraph{Fill-in-the-gap Probing.} One of the popular approaches to probe pre-trained language models is fill-in-the-gap probing, in which the researchers have constructed various probing datasets to test a model's ability on different aspects. \citet{goldberg2019assessing} found that BERT considers subject-verb agreement when performing the cloze task. 
\citet{petroni-etal-2019-language} reported that 
factual  knowledge  can  be  recovered surprisingly well from pre-trained language models. For instance, ``JDK is developed by [\underline{Oracle}]''. Similarly, we apply fill-in-the-gap to probe bridging by formulating bridging anaphora resolution as 
a \emph{of-Cloze test}.

\paragraph{Commonsense Knowledge Probing.} A lot of work has been carried out to analyze various types of commonsense knowledge encoded in transformer language models. \citet{talmor2019olmpics} constructed a set of probing datasets and test whether specific reasoning skills are captured by pre-trained language models, such as age comparison and antonym negation. \citet{da-kasai-2019-cracking} found that pre-trained BERT failed to encode some abstract attributes of objects, as well as visual
and perceptual properties that are likely to be assumed rather than mentioned.

In our work, we focus on investigating the effect of context on bridging inference using 
a well-established task on bridging resolution.
We extensively analyze the impacts of different contexts for bridging anaphora resolution. 
We found that a pre-trained BERT model achieves reasonable results for bridging anaphora resolution by using the word ``\textbf{\emph{of}}'' as the additional context. This indicates that pre-trained language models capture certain commonsense world knowledge for bridging.

\section{Methodology}
In this paper, we mainly investigate the following research questions: 
\begin{itemize}
    \item How important are the self-attention patterns of different heads for bridging anaphora resolution?
    \item Whether pre-trained LMs capture information beneficial for resolving bridging anaphora in English?
    \item How does distance between anaphor-antecedent and context influence pre-trained language models for bridging inference? 
\end{itemize}
We designed a series of experiments to answer these questions which will be detailed in the coming sections. In these experiments, we used PyTorch \cite{wolf-etal-2020-transformers} 
implementation of BERT-base-cased, BERT-large-cased, ROBERTA-base and ROBERTA-large pre-trained transformer language models with the standard number of layers, attention heads, and parameters. In the attention head-based experiments, we have limited our investigation only to the BERT-base-cased model as it is relatively smaller compared to other models and findings of this model can be generalized to other models as well.

\paragraph{Probing Dataset}
We used ISNotes \cite{markert-etal-2012-collective} dataset for all 
experiments.
We choose this corpus because it contains ``unrestricted anaphoric referential bridging'' annotations among all available English bridging corpora \cite{roesiger-etal-2018-bridging} which covers a wide range of different relations.
ISNotes contains 663 bridging anaphors but only 622 anaphors have noun phrase antecedents.\footnote{A small number of bridging antecedents in ISNotes are represented by verbs or clauses.} 
In our experiments, we only consider these 622 anaphors for investigation.
For any anaphor, the predicted antecedent is selected from the set of 
antecedent candidates.
This set is formed by considering all the mentions which occur before the anaphor. We obtained the candidate set for each anaphor by considering \q{gold mentions} annotated in ISNotes. Further, we observed that only 531 anaphors have antecedents in either previous 2 sentences from the anaphor or the first sentence of the document.
Therefore, in the experiments when 
antecedent candidates
are considered from the window of previous two sentences plus the document's first sentence, only 531 anaphors are considered. In all the experiments, accuracy is measured as the ratio between correctly linked anaphors to the total anaphors used in that particular experiment (not total 663 anaphors).

\begin{figure}[t]
\centering
\includegraphics[width=7.5cm]{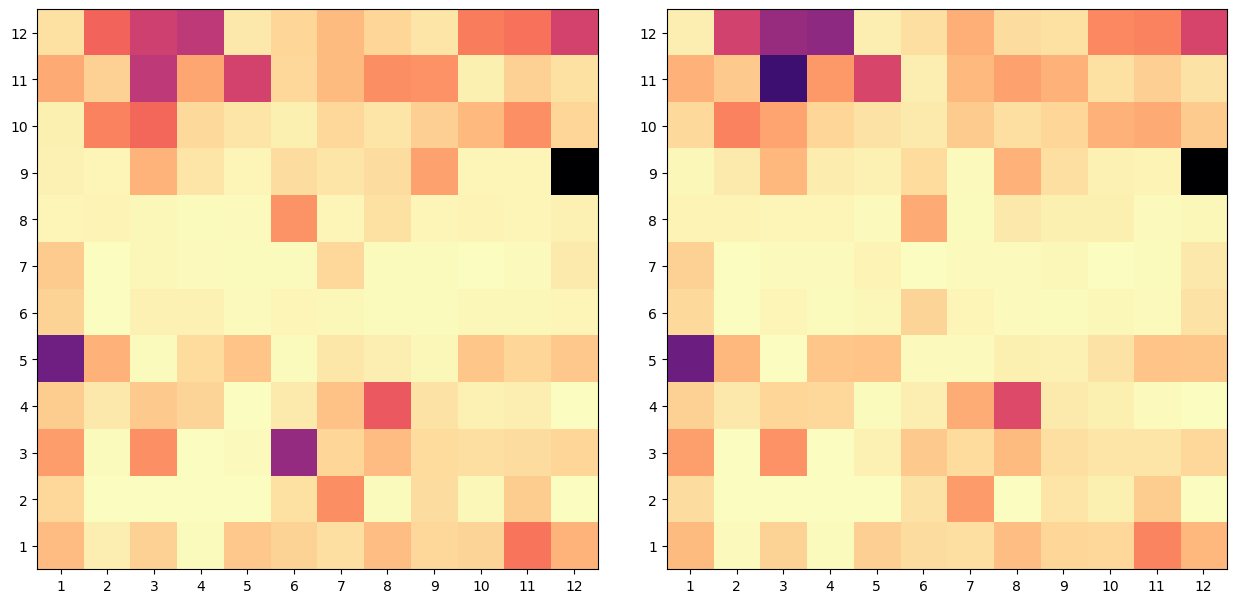}
\caption{\small Bridging signals with BERT-base-cased model with only anaphor and antecedent sentences provided. Bridging signals from anaphor to antecedent are shown in the first heatmap and the reverse signals in the second. In both heatmaps, the x-axis shows the attention head number and the y-axis shows the layer number.}\label{adj_sent}
\end{figure}

\section{Probing Individual Attention Heads}
\label{sec:head}
Attention heads are an important part of transformer based language models. Each layer consists of a certain number of attention heads depending on the model design and each attention head assigns different attention weight from every token of the input sentence to all the tokens. In our approach, we measure the attention flow between 
anaphors and antecedents for 
each attention head separately. 
In this experiment we investigate all the attention heads of every layer one-by-one. Specifically, the BERT-base-cased model used for probing contains 12 layers and 12 attention heads at each layer. Therefore, we investigate 144 attention heads for their ability to capture bridging signals.
\begin{figure}[!ht]
\centering
\begin{subfigure}[b]{0.456\textwidth}
   \includegraphics[width=1\linewidth]{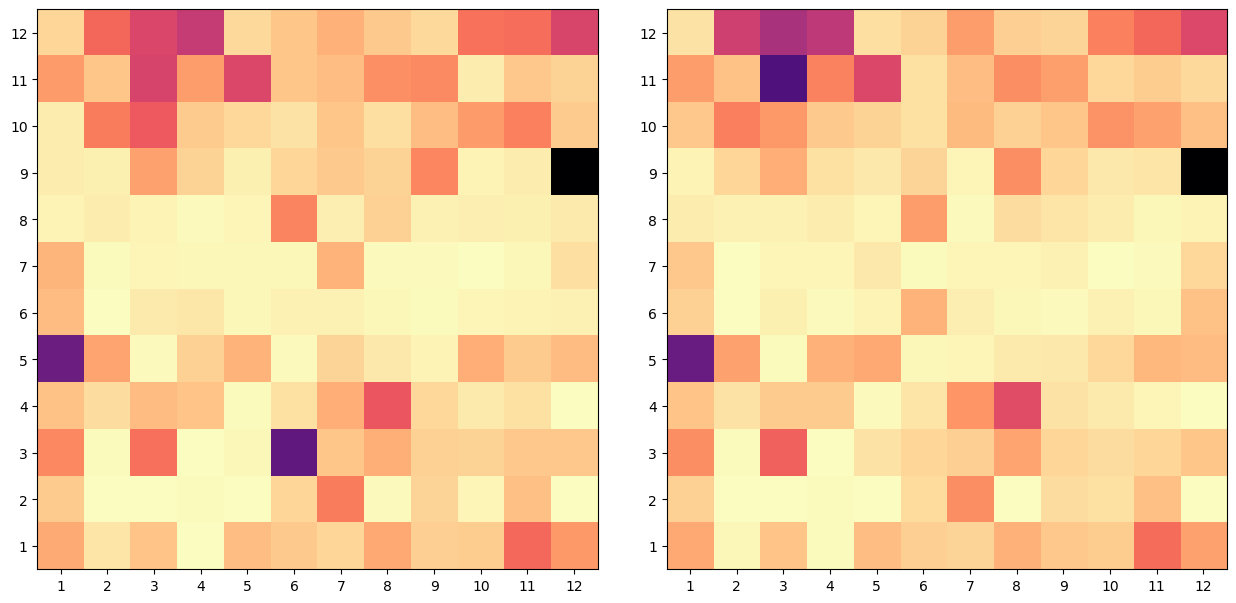}
   \caption{Anaphor-antecedent sent. distance 0}
   \label{fig:Ng1} 
\end{subfigure}

\begin{subfigure}[b]{0.457\textwidth}
   \includegraphics[width=1\linewidth]{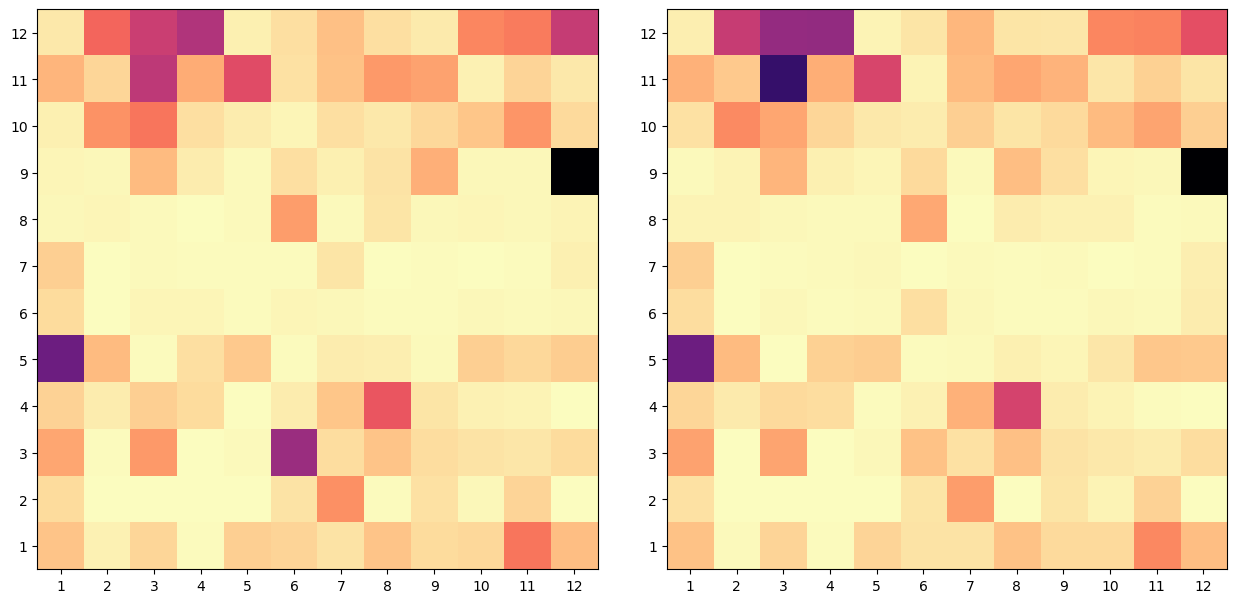}
   \caption{Anaphor-antecedent sent. distance 1}
   \label{fig:Ng2}
\end{subfigure}
\begin{subfigure}[b]{0.457\textwidth}
   \includegraphics[width=1\linewidth]{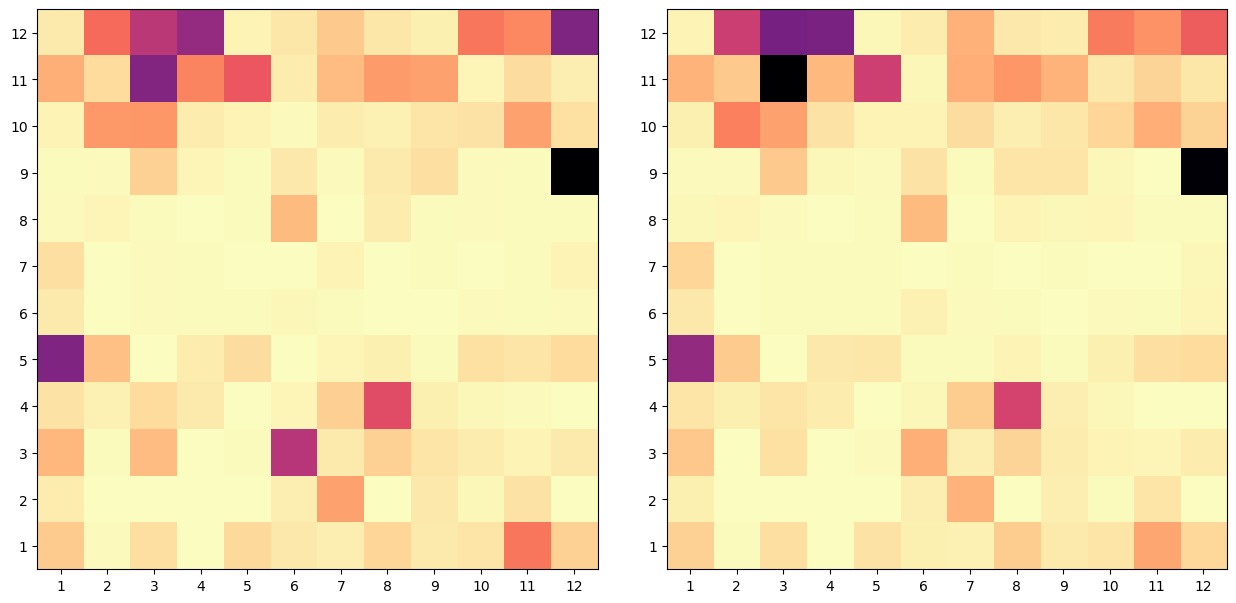}
   \caption{Anaphor-antecedent sent. distance 2}
   \label{fig:Ng3}
\end{subfigure}
\begin{subfigure}[b]{0.457\textwidth}
   \includegraphics[width=1\linewidth]{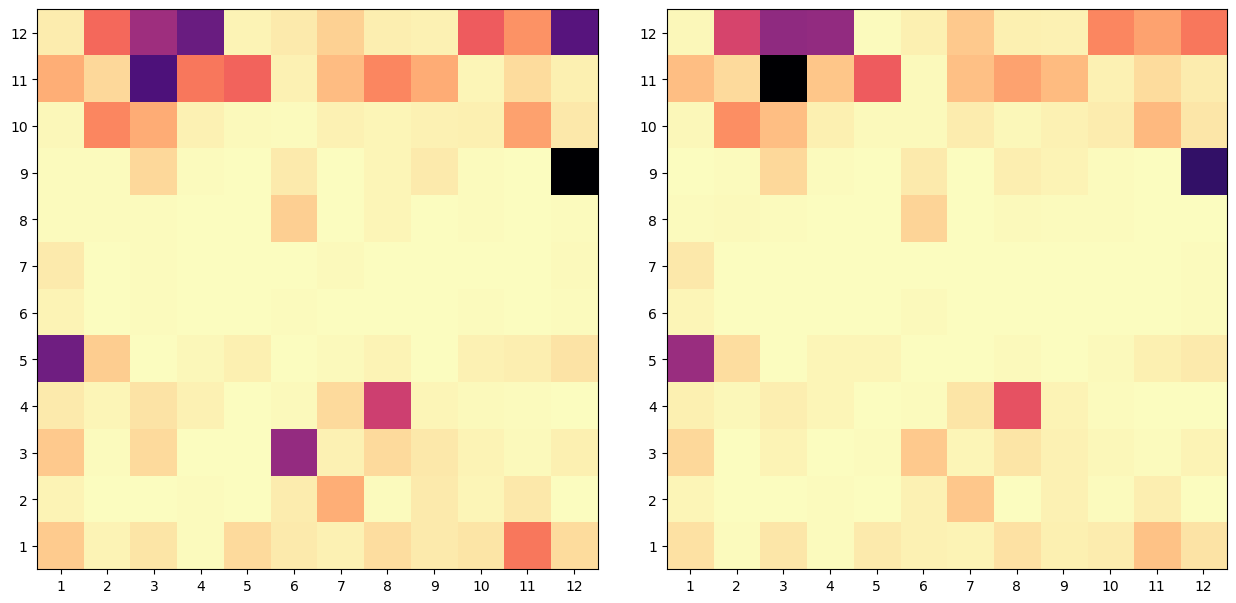}
   \caption{Anaphor-antecedent sent. distance between 3 and 5 }
   \label{fig:Ng4}
\end{subfigure}
\begin{subfigure}[b]{0.457\textwidth}
   \includegraphics[width=1\linewidth]{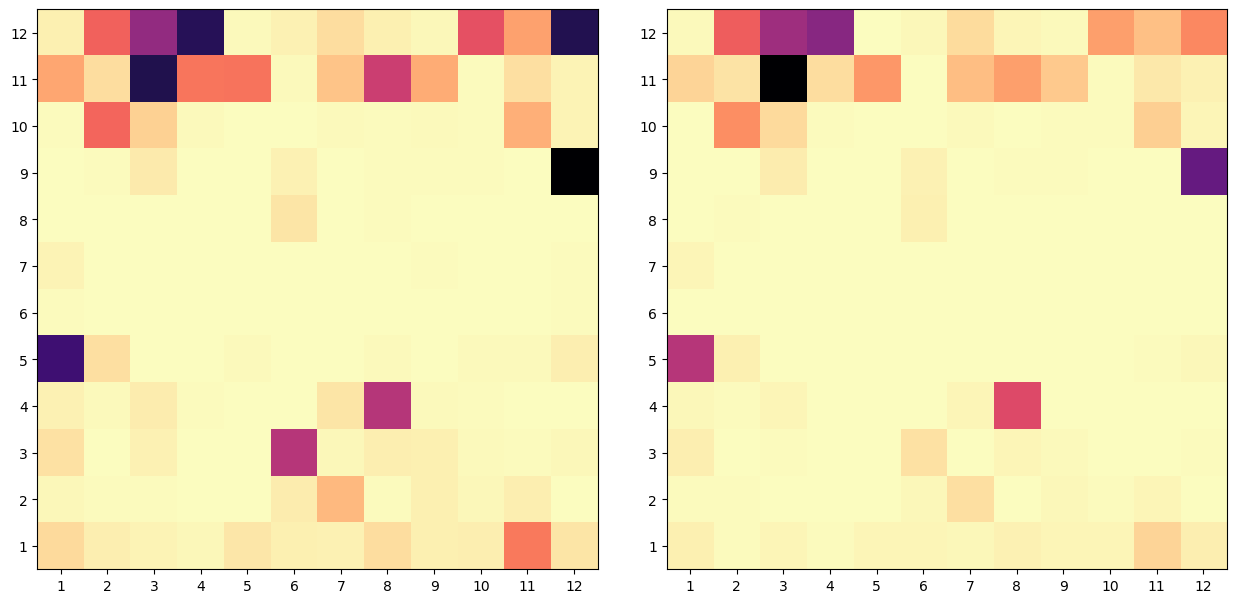}
   \caption{Anaphor-antecedent sent. distance between 6 and 10 }   \label{fig:Ng5}
\end{subfigure}
\caption{\small Bridging signals in the pre-trained BERT-base-cased model with the input including all the sentences between an anaphor and its antecedent. Different heatmaps are shown depending on the sentence distance between anaphor and antecedent. The first heatmap in each row shows the signals from anaphor to antecedent and the second one from antecedent to anaphor. All the heatmaps present the attention heads on the x-axis and the layer numbers on the y-axis. }\label{all_sent}
\end{figure}
\subsection{Bridging Signal}
\label{sec:bridgingsignal}
We look for two distinct bridging signals $-$ one from anaphor to antecedent and other from antecedent to anaphor. The bridging signal from anaphor to antecedent is calculated as the ratio of the attention weight assigned to antecedent and the total cumulative attention paid to all the words in the input. Similarly, the bridging signal from antecedent to anaphor is found in a reverse way.

There are two difficulties while getting the attention weights corresponding to anaphor or antecedent. First, the anaphor or antecedent can be a phrase with multiple words. So, we need to decide how to aggregate words' weights. 
For this, we decide to consider the semantic heads of both anaphor and antecedent, and get the attention weight between them. For instance, the semantic head for ``\emph{the political value of imposing sanction against South Africa}'' is ``\emph{value}''. Most of the time, a semantic head of an NP is its syntactic head word as in the above example.  However, for coordinated NPs such as ``\emph{the courts and the justice department}'', the syntactic head will be ``\emph{and}'' which does not reflect this NP's semantic meaning. In such cases, we use the head word of the first element as its semantic head (i.e., \emph{courts}).

Secondly, transformer language models use the wordpiece tokenizer to break words further. This produces multiple tokens from a single word if this word is absent from the language model's dictionary.  
Here, for a bridging anaphor $a$ and its head word $a_h$, we first calculate the average weight of all word piece tokens of the head word $a_h$ to other words. From these weights, we consider the weight from the anaphor $a$ to its antecedent ($w_1$). Subsequently, we add weights from $a_h$ to all other tokens present in the sentence and normalize the weight using sentence length ($w_2$). 
Note that we neglected weights assigned to special tokens (i.e. [CLS], [SEP], [PAD], etc.,) while calculating both weights as previous work suggest that these special tokens are heavily attended in deep heads and might be used as a no-op for attention heads \cite{clark-etal-2019-bert}. Finally, bridging signal is measured as the ratio between $w_1$ and $w_2$ as mentioned earlier.

\subsection{Experimental Setup}
We provide sentences containing a bridging anaphor ($Ana$) and its antecedent ($Ante$) to the pre-trained BERT model as a single sentence without the \q{[SEP]} token in-between. However, an anaphor and its antecedent do not always lie in the same or adjacent sentence(s). Therefore, we design two different experiments.
In the first setup, we provide the model with
 only those sentences which contain $Ana$ and $Ante$ while ignoring all the other sentences in-between. This setting is a bit unnatural as we are not following the original discourse narration. 
 In the second setup, we provide the model with sentences which contain $Ana$ and $Ante$ as well as all the other sentences between $Ana$ and $Ante$. 
Note that in both experiments we add markers to denote the anaphor and its antecedent 
in order to get 
exact corresponding attention weights.

\begin{table*}[!ht]
\centering
  \begin{tabular}[!]{l}
 \hline 
\rowcolor{cyan} 
\specialcell{\hspace{5cm}Easy Bridging Relations}\\
\rowcolor{LightCyan} 
\specialcell{The move will make the drug available free of charge for a time to children with \underline{the disease} and\\ symptoms of advanced \textbf{infection}.}\\
\rowcolor{Gray}
\specialcell{Last year, when the rising \underline{Orange River} threatened to swamp the course, the same engineers \\ who are pushing back the Atlantic rushed to build a wall to hold back \textbf{the flood}.} \\
\rowcolor{LightCyan}
\specialcell{At age eight, Josephine Baker was sent by her mother to \underline{a white woman's house} to do chores in \\ exchange for meals and a place to sleep -- a place in \textbf{the basement} with the coal } \\ 
\hline
\rowcolor{cyan} \specialcell{\hspace{5cm}Difficult Bridging Relations}\\
\rowcolor{Gray}
\specialcell{
In addition, \underline{Delmed}, which makes and sells a dialysis solution used in treating kidney diseases, said\\ negotiations about pricing had collapsed between it and a major distributor, National Medical Care\\ Inc. Delmed said Robert S. Ehrlich resigned as chairman, president and chief executive.\\ Mr. Ehrlich will continue as a director and \textbf{a consultant}.} \\
\rowcolor{LightCyan}
 \specialcell{The night the Germans occupied all of France, \underline{Baker} performed in Casablanca. \\ The Free French wore black arm bands, and when she sang \q{J'ai deux amours} they wept. \\ Ms.Rose is best on \textbf{the early years} and World War II.} \\
 \rowcolor{Gray}
 \specialcell{In Geneva, however, they supported Iran's proposal because it would have left the Saudi percentage\\ of \underline{the OPEC} total intact, and increased actual Saudi volume to nearly 5.3M barrels daily from 5M.\\ Some of the proposed modifications since, however, call on Saudi Arabia to \q{give back} to the\\ production-sharing \textbf{pool} a token 23,000 barrels .} \\
\hline
  \end{tabular}%
  \caption{Examples of easy and difficult bridging relations for the prominent heads to recognize. Bridging anaphors are typed in boldface, antecedents in underscore.
  }\label{prom_heads_qualt}
\end{table*}

\subsection{Results With Only Ana-Ante Sentences}
\label{sec:ana-anteSent}
 For the input of only sentences containing anaphors and antecedents, we plot the bridging signals corresponding to each attention head separately (see the heatmaps in Figure \ref{adj_sent}).
 The left heatmap shows the signals from anaphors to antecedents and the right one shows the signals from antecedents to anaphors. Both heatmaps are based on the pre-trained BERT-base-cased model. The x-axis 
represents the number of attention heads from 1-12 and the y-axis 
represents the number of layers from 1-12. The darker shade of the color indicates stronger bridging signals and brighter color indicates a weak signal. 

 The plot shows that the lower layers capture stronger bridging signal in comparison with the middle layers with an exception 
 at the first attention head in the fifth layer.
 Also, the higher layers pay most attention to bridging relations in comparison to the middle and lower layers. The observation is consistent in both 
 directions $-$ from anaphors to antecedents and from antecedents to anaphors.  
\subsection{Results With All Sentences}
As stated earlier, for 
an anaphor, the antecedent can lie in the same sentence or any previous sentence. This demands a separate investigation of bridging signals depending on the distance (measured in terms of sentences) between anaphors and antecedents. Therefore, we plot bridging signals captured by all attention heads depending on the distance between anaphors and  antecedents in Figure \ref{all_sent}. 

The first plot shows the signals between anaphors and antecedents where the distance between them 
is 0 (i.e., they occur in the same sentence).  The second and the third plots show the bridging signals between anaphors and antecedents in which the anaphor-antecedent sentence distance is 1 and 2, respectively.

In ISNotes, 77\% of anaphors have antecedents occurring in the same or up to two sentences prior to the anaphor. The remaining anaphors have distant antecedents and each distance group only contains a small number of anaphor-antecedent pairs. Therefore, we divide the remaining anaphors into two coarse groups.
The plots in Figure \ref{fig:Ng4} and Figure \ref{fig:Ng5} are plotted by combining anaphor-antecedent pairs which are apart by 3 to 5 sentences and 6 to 10 sentences, respectively. 
Note that we could not plot attention signals for bridging pairs with sentence distance longer than 10 sentences because of the limitation of the input size in BERT. 

We observe that, the patterns which are visible with only anaphor-antecedent sentences as the input (Section \ref{sec:ana-anteSent}) are consistent even with considering all the sentences between anaphors and antecedents. It is clear that higher layers attend more to bridging relations in comparison with lower  and middle layers. Also, the lower layers fail to capture bridging signal as the distance between anaphors and   antecedents increases. Attention weights assigned by 
certain attention heads 
(5:1, 9:12, 11:3 and 12:2-4) are fairly consistent. One more important thing to observe is that as the distance between anaphors and antecedents increases the overall bridging signal decreases. This can be observed by looking at all the heatmaps in Figure \ref{all_sent} as the heatmaps with lower distances are on the darker side.

\subsection{Discussion}
Based on the results from the previous two experiments, we observed that in the pre-trained BERT model, the higher layers pay more attention to bridging relations in comparison with the middle and the lower layers.
This observation is in-line with other studies in which 
the authors found that
simple surface features were captured in the lower layers and complex phenomenons like coreference were captured in the higher layers \cite{jawahar-etal-2019-bert}. Also, the overall attention decreases with the increase in the distance between anaphors and antecedents.

We also observed that there are some prominent attention heads which consistently capture bridging relations (5:1, 9:12, 11:3 and 12:2-4). 
In order to check which bridging relations are easier or harder for these prominent attention heads to capture, we further investigated qualitatively to identify bridging pairs that get higher or lower attentions in these attention heads. Specifically, we consider pairs which have the bridging 
signal ratio (defined in Section \ref{sec:bridgingsignal})
more than 70\% as easier bridging relations for BERT heads to recognize. If the bridging signal ratio is less than 10\%,  then the corresponding bridging relation is considered as difficult for BERT heads to identify. We list a few easy and difficult examples in Table \ref{prom_heads_qualt}. In general, we observe that semantically closer pairs are easy for prominent heads to identify (e.g., house-basement, disease-infection). On the other hand, pairs that are distant and require more context-dependent   as well as common-sense knowledge inference are difficult for the prominent heads to recognize. 



\section{Fill-in-the-gap Probing: LMs as Bridging Anaphora Resolvers}
\label{sec:fill-in-the-gap}
The transformer-based language models are trained with an objective to predict the masked tokens given the surrounding context. Thus, they can also produce a score for a word which can be placed at the masked token in a given sentence. We make use of this property of the language models and propose a novel formulation to understand the bridging anaphora resolution capacity of the pre-trained language models.

\subsection{Of-Cloze Test}
The syntactic prepositional structure (\emph{X of Y}, such as \q{the door of house} or \q{the chairman of company}) encodes a variety of bridging relations. Previous work has used this property to design features and develop embedding resources for bridging \cite{hou-etal-2013-global,hou-2018-deterministic,hou-2018-enhanced}.

Inspired by this observation, we formulate bridging anaphora resolution as a cloze task. Specifically, given a bridging anaphor and its context, we insert ``of [MASK]'' after the head word of the anaphor (see Example \ref{ex:bridging2}). We then calculate the probability of each candidate to be filled as the mask token. The highest scoring candidate is selected as the predicted antecedent for the anaphor. One of the advantages of our formulation is that we can easily control the scope of the context for each bridging anaphor (e.g., \emph{no-context}, \emph{local context} or \emph{global context}). This allows us to test the effect of different types of context for bridging inference. 

\begin{examples}
\item \label{ex:bridging2} \emph{Original context}: The survey found that over a three-year period 22\% of \underline{the firms} said employees or owners had been robbed on their way to or from work or while on the job. \textbf{Seventeen percent} reported their customers being robbed.
\\[1mm] 
\emph{Cloze test context}: The survey found that over a three-year period 22\% of \underline{the firms} said employees or owners had been robbed on their way to or from work or while on the job. \textbf{Seventeen percent of [MASK]} reported their customers being robbed. 
\end{examples}
\subsection{Experimental Setup}

Recall that in our \textit{Of-Cloze test}, 
antecedent candidates
are provided and the highest scoring candidate is selected as the predicted antecedent. These candidates are formed by considering mentions which are occuring prior to the anaphor. We design two different experiment sets based on the scope of 
antecedent candidates and the 
surrounding context.

\paragraph{Candidates Scope}
In the first set of experiments, we consider two different sets of 
antecedent candidates
for an anaphor $a$. The first set contains \emph{salient and nearby mentions} 
as 
antecedent candidates.
Here, mentions only from the first sentence of the document, previous two sentences preceding $a$ and the sentence containing $a$ are considered as candidates. 
This setup follows previous work on selecting antecedent candidates \cite{hou-2020-bridging}.
The second set contains \textit{all mentions} occurring before the anaphor $a$ from the whole document. The second setup of forming antecedent candidates is more challenging than the first one because the number of candidates increases which makes selecting the correct antecedent difficult.

Next, we provide the same context for anaphors in both of the experiments described above. We construct the context $c$ for the bridging anaphor $a$. Precisely, $c$ contains the first sentence of the document, the previous two sentences occurring before $a$, as well as the sentence containing $a$. We replace the head of $a$ as ``of [MASK]''.


We also compare this fill-in-the-gap probing approach with the attention heads-based approach for resolving bridging anaphors. Specifically, we use the prominent heads in BERT for identifying bridging relations from 
Section \ref{sec:head}.
Here, we obtained attention weights from an anaphor head to all antecedent candidate heads by adding attentions from prominent heads 5:1, 9:12, 11:3, and 12:2-4. Then the highest scoring candidate is predicted as the antecedent for the anaphor. 

\paragraph{Context Scope}
In the second set of experiments, we concentrate on probing the behavior of language models at capturing bridging relations with different contexts. 
We experiment with the following four settings: 
\begin{itemize}
    \item a. Only anaphor: in this setup, only the anaphor phrase (with \q{of [MASK]} being inserted after the anaphor's head word) is given as the input to the model.
    \item b. Anaphor sentence: the sentence containing the anaphor is provided. The phrase \q{of [MASK]} is inserted after the head word of the anaphor.
    \item c. Ante+Ana sentence: on top of b, the sentence containing the antecedent  is also included in the context.
    \item d. More context: on top of b, the first sentence from the document as well as the previous two sentences preceding the anaphor are included.
   
\end{itemize}
\paragraph{Without ``of'' Context} 
  To test the effect of the strong bridging indicating signal  ``\emph{of}'', we further execute another set of experiments. Specifically, We remove \q{of} from \q{anaphor$_{head}$ of [MASK]} and instead, provide \q{anaphor$_{head}$ [MASK]} for each type of the context described above. 
  
  \paragraph{Perturbed Context}
  In this setting, we perturb the context by randomly shuffling the words in the context except for the anaphor and antecedent phrases for each type of the context mentioned above. Note that we still have the ``\emph{of}'' indicator in this setup.

\begin{table*}
\centering
  \begin{tabular}[h]{|l|c|c|c|c|c|}
    \hline
 \specialcell{\textbf{Antecedent} \\ \textbf{Candidate Scope}} & \specialcell{\textbf{No.}\\ \textbf{Anaphors}} &\specialcell{\textbf{BERT-} \\ \textbf{Base}} & \specialcell{\textbf{BERT-}\\ \textbf{Large}} & \specialcell{\textbf{RoBERTa-} \\ \textbf{Base}} & \specialcell{\textbf{RoBERTa-} \\ \textbf{Large}} \\\hline
 \multicolumn{6}{|c|}{\emph{Prominent attention heads}} \\ \hline
 \specialcell{(1) Salient/nearby mentions} & 531 & 20.15 &-&-&-\\ \hline
 \multicolumn{6}{|c|}{\emph{Of-Cloze Test}} \\
\hline
\specialcell{(2) Salient/nearby mentions} & 531 & 31.64 & 33.71 & 34.08 & \textbf{34.65} \\ \hline
 \specialcell{(3) All previous mentions} & 622 & 26.36 & 28.78 & 27.49 & \textbf{29.90} \\ \hline
 \multicolumn{6}{|c|}{\emph{Of-Cloze Test: Anaphors with antecedents in the provided contexts}} \\ \hline
  \specialcell{(4) All previous mentions 
 }  & 531 & 29.00 & 30.88 & 30.32 & \textbf{32.39} \\ \hline
 \multicolumn{6}{|c|}{\emph{Of-Cloze Test: Anaphors with antecedents outside of the provided contexts}} \\ \hline
 \specialcell{(5) All previous mentions 
} & 91 & 10.98 & \textbf{16.48} & 10.98 & 15.38 \\ \hline
  \end{tabular}%
\caption{Result of selecting antecedents for anaphors with two different probing approaches (\emph{Prominent attention heads} and \emph{Of-Cloze Test}) based on the same context. Accuracy is calculated over a different number of anaphors.
}\label{lm_res}
\end{table*}

\begin{table}
\centering
  \begin{tabular}[h]{cc}
    \toprule
\textbf{Distance} & \textbf{Accuracy} \\\midrule
salient$^*$ & 38.65 \\
0 & 26.92\\
1 & 20.58\\
2 & 17.30\\
>2 & 10.98\\\bottomrule
  \end{tabular}%
  \caption{Anaphor-antecedent distance-wise accuracy with the  BERT-base-cased model. 
  $^*$ indicates that the antecedent is in the first sentence of the document.}\label{dist_acc}
\end{table}

\subsection{Results and Discussion}
\subsubsection{Results on Candidates Scope}
Table \ref{lm_res} shows the accuracy of using only the
\textit{prominent heads} and our \emph{Of-Cloze} test approach for bridging anaphora resolution. All experiments are based on the same context (i.e., the sentence containing an anaphor, the previous two sentences preceding the anaphor as well as the first sentence from the document).

We find that the \emph{Of-Cloze} probing approach achieves higher result 
in comparison to the prominent attention head approach (31.64\% vs. 20.15\%)
under the same conditions.
One reason might be that although other attention heads do not significantly attend to bridging relations but cumulatively they are effective. 

We also observe that in the \emph{Of-Cloze} test, the results of  using salient/nearby mentions as antecedent candidates are better than 
choosing antecedents from all previous mentions (Row (2) vs. Row (3), and Row (2) vs. Row (4)). This is because the model has to choose from a smaller number of candidates in the first case as the average number of antecedent candidates are only 22 per anaphor as opposed to 148 in the later case.


We further divide 622 anaphors in Row (3) into two groups (Row (4) and Row (5) in Table \ref{lm_res}) depending on whether the corresponding antecedents occur in the provided contexts. It can be seen that the performance is significantly better when antecedents occur in the contexts. 


Finally, when comparing the results of each language model in each row separately, it seems that the bigger models are always better at capturing bridging information. In general, the RoBERTa-large model performs better than other models except when antecedents do not occur in the provided contexts (Row (5)).


Note that the results in Table \ref{lm_res} are not calculated over all 663 anaphors in ISNotes. Therefore, if the results are normalized over all anaphors then we get the best result with the RoBERTa-large model (28.05\%), which is reasonably fine in comparison with the state-of-the-art result of 50.08\% \cite{hou-2020-bridging} given that the model is not fine-tuned for the bridging task. 

\subsubsection{Results on Ana-Ante Distance}
We further analyze the results of choosing antecedents obtained using the BERT-base-cased model with all previous mentions as the antecedent candidate scope in our Of-Cloze test probing experiment (Row (3) in Table \ref{lm_res}) to understand the effect of distance between anaphors and antecedents. 
The results are shown in Table \ref{dist_acc}.

In general, it seems that the accuracy decreases as the distance between anaphors and antecedents increases except when antecedents are from the first sentences of the documents. This is related to the position bias in news articles from ISNotes. Normally globally salient entities are often introduced in the beginning of a new article and these entities are preferred as antecedents. 

The other reason for the lower results in case of antecedents being away for more than two sentences might be that these antecedents are absent from the provided context.


\begin{table}
\centering
  \begin{tabular}[h]{llll}
    \toprule
\textbf{Context Scope} &
\specialcell{\textbf{with} \\ \textbf{\q{of}}} & \specialcell{\textbf{without} \\ \textbf{\q{of}}} & \textbf{perturb}\\\midrule
only anaphor & 17.20 & 5.62 & -\\
ana sent. & 22.82 & 7.71 & 10.28 \\
ana+ante sent. &27.81 & 9.61 & 10.93\\
more context & 26.36 & 12.21 & 11.41 
\\\bottomrule
  \end{tabular}%
  \caption{Accuracy of selecting antecedents with different types of context using BERT-of-Cloze Test.}\label{cont_imp}
\end{table}
\subsubsection{Results on Different Contexts}

The results of experiments with different types of context are shown in Table \ref{cont_imp}. All experiments are based on the  BERT-base-cased model with all previous mentions as the antecedent candidate scope. We refer to this model as \emph{BERT-Of-Cloze} in the following discussion.

In the first column of the table, 
\emph{BERT-Of-Cloze} achieves an accuracy score of 17.20\% with only the anaphor information plus ``\emph{of [mask]}''.
We can see that the results improve incrementally with the addition of context. More specifically, the accuracy score improves from 17.20\% to 22.82\% by adding sentences containing anaphors. Adding sentences which contain antecedents (\emph{ana + ante sent.}) further improves the accuracy score to 27.81\%. Finally, adding more local context and the first sentence leads to an accuracy score of 26.36\%.
Note that compared to ``\emph{ana + ante sent.}'', ``\emph{more context}'' represents a more realistic scenario in which we do not assume that the antecedent position information is known beforehand.    
In general, 
the results in the first column of Table \ref{cont_imp} indicate
that the model can leverage context information when predicting antecedents for bridging anaphors.


Results reduce drastically when \emph{\q{of}} is removed from the ``anaphor of [MASK]'' phrase (Table \ref{cont_imp}, column:2) from all context scopes. Without this indicator, the language model cannot make sense of two adjacent tokens such as ``\emph{consultant company}''.

It is interesting to see that the results reduced drastically as well when we perturb the context between the anaphor and antecedent (Table \ref{cont_imp}, last column). This establishes the importance of meaningful context for performing bridging inference effectively in transformer language models. 
\subsection{Error Analysis: \emph{Of-Cloze test}}
We analyzed anaphor-antecedent pairs that are linked wrongly by the \emph{Of-Cloze}
formulation and observed some common erros.




\paragraph{Failure at capturing sophisticated common-sense knowledge: } We found that the pre-trained transformer language model such as BERT acquires simple common-sense knowledge, therefore it can link anaphor-antecedent pairs such as ``\textit{sand--dunes}'' and ``\textit{principal--school}''. But it fails at capturing sophisticated knowledge, such as  ``\textit{consultant--Delmed (a company)}'' and ``\textit{pool--OPEC (Organization of petroleum countries)}''. This might be happening because of the rare co-occurrences of these pairs in the original text on which BERT is pre-trained. Also, BERT has inherent limitations at acquiring such  structured knowledge \cite{9054147}.

\paragraph{Language modelling bias: } In our \emph{Of-Cloze test} probing,  we use pre-trained transformer language models without fine-tuning. 
As a result, the model fills masked tokens that are fit according to the language modeling objective, not for bridging resolution. Thus, sometimes, the selected token perfectly makes sense in the single sentence but the choice is incorrect in the broader context.
Consider the example, \q{Only 22\% of [MASK] supported private security patrols [...]}. BERT predicts ``\textit{police}'' as a suitable antecedent that produces a meaningful local sentence. However, the correct antecedent is ``\textit{correspondents}'' according to the surrounding context of this sentence.
\paragraph{Unsuitable formulation for set-relations: } Our \emph{Of-Cloze} formulation produces awkward phrases for some bridging pairs that possess set-relations. Considering a bridging pair $-$ ``\emph{One man - employees}'', in this case the model should assign high score for the phrase $-$ \q{One man \textit{of} employees}. But, as this phrase is quite clumsy, BERT naturally being a language model assigns low scores for these pairs.

\section{Conclusions}
We investigated the effectiveness of pre-trained transformer language models in capturing bridging relation inference by employing two distinct but complementary approaches. 

In the first approach, we probed individual attention heads in BERT and observed that attention heads from higher layers prominently captured bridging compared to the middle and lower layers and some specific attention heads consistently looked for bridging relation. In our second approach, we considered using language models for bridging anaphora resolution by formulating the task as a \emph{Of-Cloze} test.
We carefully designed experiments to test the influence of different types of context for language models to resolve  bridging anaphors. 
Our results indicate that pre-trained transformer language models encode substantial information about bridging.



Finally, in this work, we only focus on understanding the capacity of the pre-trained language models for bridging inference. Based on  the insights we gained from the current probing study, in the future, we plan to explore how to better use pre-trained transformer language models for bridging resolution. 


\section*{Acknowledgements}
We thank the three anonymous reviewers for their comments and feedback. This work was partially supported by the French National Research
Agency via grant no ANR-16-CE33-0011-01 as well as by CPER Nord-Pas de Calais/FEDER DATA Advanced data science and technologies 2015-2020.

\bibliography{anthology,custom}
\bibliographystyle{acl_natbib}

\appendix


\end{document}